\pdfoutput=1
\documentclass{article}

\usepackage{microtype}
\usepackage{graphicx}
\usepackage{subfigure}
\usepackage{caption}
\usepackage{float}
\usepackage{algorithm}
\usepackage[noend]{algpseudocode}
\usepackage{relsize}
\usepackage{lipsum}

\usepackage{booktabs} 
\usepackage{amsmath}
\usepackage{amssymb}
\usepackage{fixltx2e}
\usepackage[shortlabels]{enumitem}
\usepackage{enumitem}
\usepackage{mathtools}

\setlist{nolistsep}

\newtheorem{definition}{Definition}
\newtheorem{theorem}{Theorem}
\newtheorem{corollary}{Corollary}

\newtheorem{conjecture}{Conjecture}
\usepackage{enumitem}

\usepackage{hyperref}

\usepackage[accepted]{icml2020}

\begin{document}
\allowdisplaybreaks
\setlength{\abovedisplayskip}{0pt}
\setlength{\belowdisplayskip}{0pt}
\setlength{\abovedisplayshortskip}{0pt}
\setlength{\belowdisplayshortskip}{0pt}
\setlength{\topsep}{0pt}

\twocolumn[
\icmltitle{A Causal Linear Model to Quantify Edge Flow and Edge Unfairness for Unfair Edge Prioritization \& Discrimination Removal}



\icmlsetsymbol{equal}{*}

\begin{icmlauthorlist}
\icmlauthor{Pavan Ravishankar}{equal,to,iitm}
\icmlauthor{Pranshu Malviya}{equal,to,iitm}
\icmlauthor{Balaraman Ravindran}{to,iitm}
\end{icmlauthorlist}

\icmlaffiliation{to}{Robert Bosch Centre for Data Science and Artificial Intelligence, IIT Madras, Chennai, India}
\icmlaffiliation{iitm}{Department of Computer Science and Engineering, IIT Madras, Chennai, India}
\icmlcorrespondingauthor{Pavan Ravishankar}{cs17s026@smail.iitm.ac.in}
\icmlcorrespondingauthor{Pranshu Malviya}{cs19s031@cse.iitm.ac.in}
\icmlcorrespondingauthor{Balaraman Ravindran}{ravi@cse.iitm.ac.in}

\icmlkeywords{Machine Learning, ICML}

\vskip 0.3in
]



\printAffiliationsAndNotice{\icmlEqualContribution} 

\begin{abstract}
Law enforcement must prioritize sources of unfairness before mitigating their underlying unfairness, considering that they have limited resources. Unlike previous works that only make cautionary claims of discrimination and de-biases data after its generation, this paper attempts to prioritize unfair sources before mitigating their unfairness in the real-world. We assume that a causal bayesian network, representative of the data generation procedure, along with the sensitive nodes, that result in unfairness, are given. We quantify \textit{Edge Flow}, which is the belief flowing along an edge by attenuating the indirect path influences and use it to quantify \textit{Edge Unfairness}. We \textit{prove} that cumulative unfairness is non-existent in any decision, like judicial bail, towards any sensitive groups, like race, when the edge unfairness is absent, given an error-free linear model of conditional probability. We then measure the \textit{potential} to mitigate the cumulative unfairness when edge unfairness is decreased. Based on these measures, we propose an \textit{unfair edge prioritization} algorithm that prioritizes the unfair edges and a \textit{discrimination removal procedure} that de-biases the generated data distribution. The experimental section validates the specifications used for quantifying the above measures.
\end{abstract}
\section{INTRODUCTION}
\textbf{Motivation and Problem:} Anti-discrimination laws prohibit unfair treatment of people based on sensitive nodes, such as gender (\cite{act1964civil}). The fairness of a decision process is based on \textit{disparate treatment} and \textit{disparate impact}. Disparate treatment, referred to as intentional discrimination, is when sensitive information is used to make decisions. Disparate impact, referred to as unintentional discrimination, is when decisions hurt certain sensitive groups even when the policies are neutral. For instance, only candidates with a height of $6$ feet and above are selected for basketball teams. Unjustifiable disparate impact is unlawful (\cite{barocas2016big}). In high-risk decisions, such as in the criminal justice system, it is imperative to identify potential sources of unfairness contributing to either disparate treatment or disparate impact and investigate the reasons for unfairness. Considering that law enforcement has limited resources, it is essential to prioritize these potential sources of unfairness before mitigating their effect. This paper proposes a methodology to prioritize potential sources of unfairness. 

\begin{figure}[t!]
\centering
\includegraphics[width=0.45\linewidth]{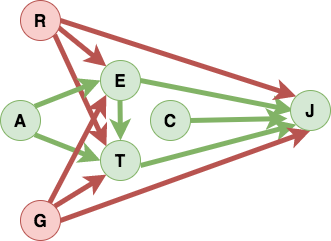}
\caption{Bail Decision Causal Graph. \textbf{Edges:} Fair Edges in Green, Unfair Edges in Red; \textbf{Nodes:} Sensitive nodes in Red, Not sensitive nodes in Green; R: Race; A: Age, G: Gender, E: Education, T: Training, C: Case characteristics, J: Judicial Bail decision} \label{main_graph}
\end{figure}

To motivate our problem, we discuss the following illustration. Consider the problem of reducing unfairness in the \textit{bail decision} towards a specific \textit{racial} group. The causal graph shown in Figure \ref{main_graph} represents the bail decision procedure with each node depending on its parents. Edges emanating from sensitive nodes, such as $R\,\to\, J$, are designated as unfair and are potential sources of unfairness (\cite{chiappa2018causal}). The racial group $R$ affects bail decision $J$ via paths from $R$ to $J$. Even though discrimination towards racial group $R$ in bail decision $J$ can be quantified (\cite{zhang2017causal}), it only serves as a cautionary claim lacking information required to mitigate the effect. 

This paper attempts to provide tangible information by quantifying a few terms. \textit{First}, the belief flowing only along the directed edge $R \rightarrow J$ is designated as \textit{edge flow}. The influences along other paths from $R$ to $J$ are attenuated by the term designated as the \textit{scaling factor}. \textit{Second}, we quantify \textit{edge unfairness} which is the difference in the value of the conditional probability table $CPT$ of $J$ with and without edge flow along $R \rightarrow J$. \textit{Third,} we measure the \textit{potential to mitigate cumulative unfairness} in a decision $J=j$ towards group $R=r$ when edge unfairness in $R \rightarrow J$ is decreased. Cumulative unfairness captures discrimination due to unequal influences of $R=r$ on $J=j$ as compared to a different $R=r'$ on $J=j$ via the directed paths from $R$ to $J$. 

Further, this paper uses the aforementioned quantities to construct the following algorithms. \textit{First,} by proving a theorem that shows that cumulative unfairness is non-existent when edge unfairness is non-existent, we propose an \textit{unfair edge prioritization algorithm} using edge unfairness and the potential to mitigate cumulative unfairness. For instance, let (0.3, 0.4) in $R \rightarrow J$, (0.2, 0.3) in $R \rightarrow L$, and (0.1, 0.2) in $R \rightarrow E$ be edge unfairness and the potential to mitigate cumulative unfairness, respectively. Using additive sum, priorities of 0.7, 0.5, and 0.3 are assigned to edges. Using these priorities, law enforcement can then address the real-world issues underlying the highest priority edge $R \rightarrow J$ such as lack of representation of $R=r$ in the judicial panel. \textit{Fifth,} we propose a \textit{discrimination removal algorithm} to de-bias data by formulating an optimization problem with edge unfairness minimization criterion, taking the motivation from the theorem result. The set of discrimination constraints do not grow exponentially in the number of sensitive nodes and their values.

\textbf{Assumptions:} We assume the following. A \textit{causal bayesian network} $CBN$ consisting of a causal graph $\mathbb{G}$ to represent the decision procedure and conditional probability tables $\mathbb{P}(V|Pa(V))$, $\forall V \in \mathbb{G}$. The sensitive nodes whose emanating edges are \textit{unfair edges}. The variables in $CBN$ are discrete and observed to not deviate into the challenges of inference and identifiability (\cite{avin2005identifiability}).

\textbf{Contributions:}
\begin{enumerate}[nosep]
    \item \textit{\textbf{Quantify the Edge Flow}} by decomposing $\mathbb{P}(X|Pa(X))$ into beliefs along the edges from $Pa(X)$ to $X$.
    \item \textit{\textbf{Quantify the Edge Unfairness}} in $e=A \rightarrow X$ as the average unit contribution of edge flow in $e$ to $\mathbb{P}(X|Pa(X))$.
    \item \textit{\textbf{Prove}} that discrimination in any decision towards any sensitive groups is non-existent when edge unfairness is eliminated for a error free linear model $f$ of $CPTs$. We prove for the trivial linear case and plan to extend to the non-trivial non-parametric case in the future.
    Cumulative unfairness cannot be expressed in terms of edge unfairness in the non-parametric case. 
    \item \textit{\textbf{Quantify}} the potential to mitigate cumulative unfairness when edge unfairness is decreased. 
    \item \textit{\textbf{Propose Unfair Edge Prioritization}} algorithm to prioritize unfair edges based on their potential to mitigate cumulative unfairness and amount of edge unfairness. 
    \item \textit{\textbf{Propose Discrimination Removal algorithm}} to eliminate discrimination constraints that grow exponentially in the number of nodes and their values.
\end{enumerate}

\textbf{Contents:} We discuss preliminaries in Section 2; quantify edge flow in Section 3; quantify edge unfairness and its impact on cumulative unfairness in Section 4; prove that discrimination is absent when edge unfairness is eliminated in Section 4; propose unfair edge prioritization and discrimination removal algorithm in Section 5; discuss experiments in Section 6; discuss related work in Section 7; discuss conclusion in Section 8.

\section{PRELIMINARIES}
\label{preliminaries}
Throughout the paper, we use $\mathbf{X}$ to denote a set of nodes; $X$ to denote a single node; $\mathbf{x}$ to denote a specific value taken by $\mathbf{X}$; $x$ to denote a specific value taken by its corresponding node $X$. $\mathbf{x}_{\mathbf{A}}$ restricts the values of $\mathbf{x}$ to node set $\mathbf{A}$. $Pa(X)$ denotes parents of $X$ and $pa(X)$ the specific values taken by them. Each node is associated with a conditional probability table $\mathbb{P}(X|Pa(X))$.

\begin{definition}
\textbf{Node interventional distribution} denoted by $\mathbb{P}(\mathbf{Y}|do(\mathbf{X}=\mathbf{x}))$ is the distribution of $\mathbf{Y}$ after forcibly setting $\mathbf{X}$ to $\mathbf{x}$. $\mathbb{P}*$ is the set of all node interventional distributions $\mathbb{P}(\mathbf{Y}|do(\mathbf{X}=\mathbf{x}))$
\cite{pearl2009causality}.
\end{definition}

\begin{definition}
Let $\mathcal{G}_{\mathbf{X}}$ be graph $\mathcal{G}$ with incoming edges of $\mathbf{X}$ removed. A directed acyclic graph $\mathcal{G}(\mathbf{E,V})$ ($DAG$) with observed variables $\mathbf{V}$ is a \textbf{Causal Bayesian Network} ($CBN$) compatible with $\mathbb{P}*$ [Definition 1.3.1 \cite{pearl2009causality}, \cite{tian2003studies}] i.f.f, 
\end{definition}
\begin{enumerate}
    \itemsep0em 
    \item $\forall \mathbb{P}(\mathbf{V}|do(\mathbf{X=x})) \in \mathbb{P}*, \mathbb{P}(\mathbf{V}|do(\mathbf{X=x}))$ is Markov relative to $\mathcal{G}_{\mathbf{X}}$ which means it factorizes over $\mathcal{G}_{\mathbf{X}}$ when $\mathbf{v}$ is consistent with $\mathbf{x}$ i.e., 
    \begin{align}
    \mathbb{P}(\mathbf{v}|do(\mathbf{X=x}))=
    \underset{V \in \mathbf{V}\backslash \mathbf{X}} {\prod}\mathbb{P}(v|pa(V))\rvert_{\mathbf{X=x}}
    \end{align}
    \item $\mathbb{P}(v|pa(V),do(\mathbf{X=x}))=1, \forall V \in \mathbf{X}$ \\ ~~~[when $v$ is consistent with $\mathbf{x}$]
    \item $\mathbb{P}(v|pa(V),do(\mathbf{X=x}))=\mathbb{P}(v|pa(V)), \forall V \not\in \mathbf{X}$ \\ ~~~[when $pa(V)$ is consistent with $\mathbf{x}$]
\end{enumerate}
$DAG$ $\mathcal{G}$ represents the procedure for generating the dataset comprising of observed variables $\mathbf{V}$ with each $V$ generated using $Pa(V)$. We assume that $P(\mathbf{V}) > 0$ throughout the paper to avoid zero probability of node interventional distribution, path-specific nested counterfactual distribution (introduced later), etc. helping in comparing the effect of different interventions.  
\begin{definition}
\textbf{Identifiability:}
\label{section:iden}
Let $\mathbb{G}(\mathbf{V})$ be a Causal Bayesian Network $CBN$. A node interventional distribution $\mathbb{P}(\mathbf{Y}|do(\mathbf{X}=\mathbf{x}))$ is said to be \textit{identifiable} if it can be expressed using the observational probability $\mathbb{P}(\mathbf{V})$ (formal definition in \cite{pearl2009causality}). When $CBN$ comprises of only observed variables as in our work,

    \begin{align}
    \mathbb{P}(\mathbf{y}|do(\mathbf{x})) &=\underset{\mathbf{V\backslash\{\mathbf{X},Y\}}}{\sum}~~\underset{V \in \mathbf{V} \backslash \{\mathbf{X,Y}\},\mathbf{Y=y}}{~\prod}\mathbb{P}(v|pa(V)) \label{factorization}
    \end{align}
\end{definition}

\begin{definition}
A trail $V_1 \rightleftharpoons .....\rightleftharpoons V_n$ is said to be an \textbf{active trail} given a set of variables $\mathbf{X}$ in $\mathcal{G}$ if for every v-structure $V_i\rightarrow V_j \leftarrow V_k$ along the trail, $V_j$ or any descendent of $V_j$ is in $\mathbf{X}$ and no other node in the trail belongs to $\mathbf{X}$. 
\end{definition}
 
\begin{definition}
$\mathbf{A}$ is said to be \textbf{d-separated} from $\mathbf{B}$ given $\mathbf{C}$ in a graph $\mathcal{G}$ $(d\text{-}sep_{\mathcal{G}} (\mathbf{A};\mathbf{B}|\mathbf{C}))$ if there is no active trail from any $A \in \mathbf{A}$ to any $B \in \mathbf{B}$
\cite{pearl2009causality}, \cite{koller2009probabilistic}. If there is atleast one active trail from any $A \in \mathbf{A}$ to any $B \in \mathbf{B}$, then $\mathbf{A}$ is said to be \textbf{d-connected} from $\mathbf{B}$ given $\mathbf{C}$ in a graph $\mathcal{G}$ $(d\text{-}conn_{\mathcal{G}} (\mathbf{A};\mathbf{B}|\mathbf{C}))$ (see Fig. 1.3 \cite{pearl2009causality}).
\end{definition}

\begin{theorem}
If sets $\mathbf{X}$ and $\mathbf{Y}$ are d-separated by $\mathbf{Z}$ in a DAG $\mathcal{G(\mathbf{E,V})}$, then $\mathbf{X}$ is independent of $\mathbf{Y}$ conditional on $\mathbf{Z}$ in every distribution $\mathbb{P}$ that factorizes over $\mathcal{G}$. Conversely, if $\mathbf{X}$ and $\mathbf{Y}$ are not d-separated by $\mathbf{Z}$ in a DAG $\mathcal{G}$, then $\mathbf{X}$ and $\mathbf{Y}$ are dependent conditional on $\mathbf{Z}$ in at least one distribution $\mathbb{P}$ that factorizes over $\mathcal{G}$ (Theorem 1.2.4 in \cite{pearl2009causality})
\end{theorem}

\begin{definition}
\textbf{Total Causal Effect}
\label{section:pathspec}
$TE_{\mathbf{y}}(\mathbf{x}_2, \mathbf{x}_1)$ measures causal effect of variables $\mathbf{X}$ on decision variables $\mathbf{Y=y}$ when it is changed from $\mathbf{x}_1$ to $\mathbf{x}_2$ written as, 
\begin{align}
TE_{\mathbf{y}}(\mathbf{x}_2, \mathbf{x}_1) = \mathbb{P}(\mathbf{y}|do(\mathbf{x}_2)) - \mathbb{P}(\mathbf{y}|do(\mathbf{x}_1))
\end{align}
\end{definition}
\begin{definition}
\textbf{Path-specific effect} $SE_{\pi, \mathbf{y}}(\mathbf{x}_2,\mathbf{x}_1)$ measures effect of node $\mathbf{X}$ on decision $\mathbf{Y=y}$ when it is changed from $\mathbf{x}_1$ to $\mathbf{x}_2$ along the directed paths $\boldsymbol{\pi}$, while retaining $\mathbf{x}_1$ 
for the directed paths not in $\boldsymbol{\pi}$ i.e. $\tilde{\boldsymbol{\pi}}$ written in multiplicative scale,
\begin{align}
SE_{\pi,\mathbf{Y=y}}(\mathbf{x}_2, \mathbf{x}_1) = \frac{\mathbb{P}(\mathbf{y}|do(\mathbf{x}_2|_{\pi}, \mathbf{x}_1|_{\tilde{\boldsymbol{\pi}}}))}{ \mathbb{P}(\mathbf{y}|do(\mathbf{x}_1))}
\end{align}
\end{definition}
\begin{definition}
\textbf{Recanting Witness Criterion} for the set of paths $\boldsymbol{\pi}$ from $\mathbf{X}$ to $\mathbf{Y}$ is satisfied, if there exists a $W$ such that (1) there exists a path from a $X \in \mathbf{X}$ to $W$ which is a segment of a path in $\boldsymbol{\pi}$ (2) there exists a path from $W$ to $Y$ which is a segment of a path in $\boldsymbol{\pi}$ (3) there exists another path from $W$ to $Y$ which is not a segment of any path in $\boldsymbol{\pi}$. $W$ is called the \textbf{Recanting Witness.}
\end{definition}
\begin{theorem}
$\mathbb{P}(\mathbf{y}|do(\mathbf{x}|_{\boldsymbol{\pi}}, \mathbf{x'}|_{\tilde{\boldsymbol{\pi}}}))$ is identifiable if and only if the recanting witness criterion for $\boldsymbol{\pi}$ is not satisfied as shown in \cite{avin2005identifiability}.
\end{theorem}
\begin{corollary}
Let $\mathbf{S}_{\boldsymbol{\pi}}$ contain all children of $X \in \mathbf{X}$ where edge $X \rightarrow S$ is a segment of a path in $\boldsymbol{\pi}$. Similarly, let $\Tilde{\mathbf{S}}_{\boldsymbol{\pi}}$ contain all children of $X \in \mathbf{X}$ where edge $X \rightarrow S$ is a segment of a path not in $\boldsymbol{\pi}$. Then, $\mathbf{S}_{\boldsymbol{\pi}} \cap \Tilde{\mathbf{S}}_{\boldsymbol{\pi}} = \emptyset$ if and only if the recanting witness criterion for $\boldsymbol{\pi}$ is not satisfied. If the recanting witness criterion is not satisfied,
$\mathbb{P}(\mathbf{y}|do(\mathbf{x}|_{\boldsymbol{\pi}}, \mathbf{x'}|_{\tilde{\boldsymbol{\pi}}}))$ is given by the \textbf{\textit{edge g-formula}} which is constructed by replacing $\mathbf{x'}$ with $\mathbf{x}$ for the terms in $\mathbf{S}_{\boldsymbol{\pi}}$ in the
factorization of $\mathbb{P}(\mathbf{y}|do(\mathbf{x}))$ as shown in \cite{maathuis2018handbook}, \cite{shpitser2013counterfactual}. $\mathbb{P}(\mathbf{y}|do(\mathbf{x}|_{\boldsymbol{\pi}}, \mathbf{x'}|_{\tilde{\boldsymbol{\pi}}}))$ is,

\begin{align}
\resizebox{.43 \textwidth}{!}{$\underset{\mathbf{V} \backslash (\mathbf{X} \bigcup \mathbf{Y})}{\mathlarger{\sum}} \underset{~~V \in \mathbf{V} \backslash \{\mathbf{X,Y}\}, \mathbf{Y=y}}{\mathlarger{\prod}} \mathbb{P}(V | \mathbf{x}_{U_V}, \mathbf{x'}_{F_V}, Pa(V) \backslash \mathbf{X})$}
 \label{edgegfor}
\end{align} \label{corr1}
\end{corollary}
where $U_V$ is the set of parents of $V$ along an unfair edge and $F_V$ is the set of parents of $V$ along an unfair edge. $U_V \cap F_V=\emptyset$ when the recanting witness criterion is not satisfied. This helps in unambiguous assignment of $U_V$ to $\mathbf{x}$ and $F_V$ to $\mathbf{x'}$.

\textbf{Sensitive Node $S$:} Sensitive nodes are attributes of social relevance like race, gender, or other attributes like hair length which if used to generate data can result in discrimination.

\textbf{Unfair edge $S \rightarrow X$:} Unfair edge is a directed edge $S \rightarrow X$ in the causal graph $\mathcal{G}$ where $S$ is a sensitive node. Set of unfair edges in $\mathcal{G}$ is denoted by $\mathbf{E}^{\text{unfair}}_{\mathcal{G}}$.
For instance, in Figure \ref{main_graph}, edge $G\,\to\,E$ is a potential source of unfairness because the accused should not be denied admission based on gender. 

\textbf{Unfair paths $\boldsymbol{\pi}^{ \text{unfair}}_{\mathbf{S},Y,\mathcal{G}}$} are the set of directed paths from sensitive node $S \in \mathbf{S} \subseteq \mathbf{S}_{\mathcal{G}}$ to the decision variable $Y$ in graph $\mathcal{G}$. \label{unfairp}
Unfair paths capture how unfairness propagates from the sensitive nodes onto a destination node. For instance, in Figure \ref{main_graph}, $\boldsymbol{\pi}^{\text{unfair}}_ {G,J,\mathcal{G}}$ consists of $G\,\to\,E\,\to\,J$ because gender $G$ is a sensitive attribute and $G\,\to\,E$ is unfair. It captures how unfairness in the edge $G\,\to\,E$ propagates to $J$. Only the causal paths from the sensitive attributes are considered to be unfair paths because the non-causal paths do not propagate the unfairness from the sensitive attribute. For instance, there is another variable, say religious belief $R$, and another non-causal path, say $R\,\leftarrow\,E\,\to\,J$ that indicates that the education $E$ decides the religious belief $R$ and bail decision $J$, in Figure \ref{main_graph}. Yet, $R\,\leftarrow\,E\,\to\,J$ is fair because bail decision $J$ is taken based on the education $E$ \cite{chiappa2018causal}.

\section{Edge Flow} \label{section4}
In this section, we quantify the belief flowing \textit{only} along a directed edge, say $R \rightarrow J$, designated as \textit{edge flow}. We decompose $\mathbb{P}(J|Pa(J))$ into beliefs flowing along the edges from $Pa(J)$ to $J$ that includes $R \rightarrow J$. To attenuate other influences from $R$ to $J$, we formulate a quantity designated as the \textit{scaling factor}. 

\begin{figure}[h]
\centering
\includegraphics[height=5.3cm,width=\linewidth]{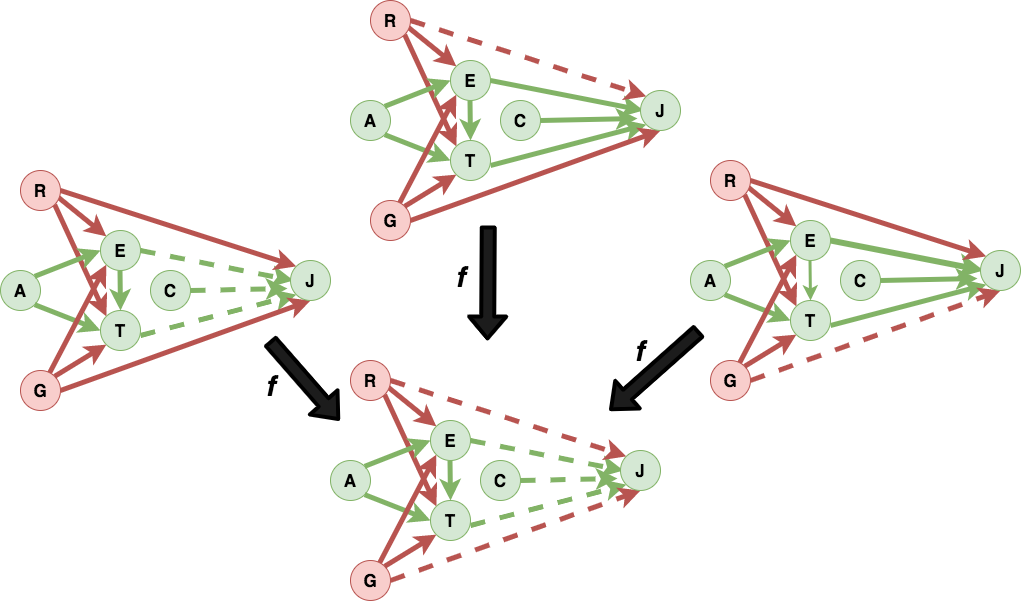}
\caption{Edge flows $\mathbb{P}^{\{E,C,T\}}_{\text{flow}}(J)$, $\mathbb{P}^{R}_{\text{flow}}(J)$ and $\mathbb{P}^{G}_{\text{flow}}(J)$ are inputs to function $f$ that models $\mathbb{P}(J|Pa(J))$. Influences along light edges are weakened due to the scaling factor.}
\label{modelgraph}
\end{figure}

Before quantifying edge flow, let us discuss the methodology of decomposing $\mathbb{P}(J|Pa(J))$ into beliefs along the edges from $Pa(J)$ to $J$ in the causal graph shown in Figure \ref{main_graph}. Let $D^{\mathbb{P}}_ {J,Pa(J)}=\{(J \not\perp X | Pa(J)\backslash X) | X \in Pa(J)\}$ be the set of possible dependencies in $\mathbb{P}(J|Pa(J))$. By Theorem \ref{dseptheorem}, active trails $\{X \rightarrow J | X \in Pa(J)\}$ are the consequences due to the dependencies in $D^{\mathbb{P}}_{J,Pa(J)}$ as shown in Figure \ref{modelgraph}. To decompose $\mathbb{P}(J|Pa(J))$ into beliefs along the parental edges of $J$, we construct beliefs known as edge flows whose possible dependencies collectively result in the same set of active trails as that of 
$\mathbb{P}(J|Pa(J))$. 

So, we construct edge flows $\mathbb{P}^{\{E,C,T\}}_{\text{flow}}(J)$, $\mathbb{P}^{R}_{\text{flow}}(J)$, and $\mathbb{P}^{G}_{\text{flow}}(J)$ each of whose possible dependencies result in active trails along the fair edges \{$E \rightarrow J$, $C \rightarrow J$, $T \rightarrow J$\}, the unfair edge \{$R \rightarrow J$\}, and the unfair edge \{$G \rightarrow J$\} respectively, as shown in Figure \ref{modelgraph}. (explained after The Scaling Factor definition). Henceforth, the edge flows can be considered as interacting via a function $f$ to generate $\mathbb{P}(J|Pa(J))$. In other words, edge flows are inputs to $f$ that models $\mathbb{P}(J|Pa(J))$ as shown in Figure \ref{modelgraph}. 

We begin by defining the \textit{scaling factor} to understand how we arrived that the active trails are consequences of the possible dependencies in an edge flow.
\begin{definition}\label{scalingfactor}
The scaling factor $S_{\mathbf{M=m} \rightarrow X=x}$ is defined as,
\begin{align}
S_{\mathbf{M=m} \rightarrow X=x} 
&=\frac{1}{|\mathbf{M} \backslash \mathbf{m}|} \underset{\mathbf{m'}\in\mathbf{M}\backslash\mathbf{m}}{\sum} {SE_{\boldsymbol{\pi},x}(\mathbf{m, m'})} \nonumber \\
\text{where}, ~\boldsymbol{\pi}&=\{M \rightarrow X|M \in \mathbf{M}, \mathbf{M} \in Pa(X)\} \label{eq:scalfac}
\end{align}
\end{definition}
$S^{\mathbb{P}}_{\mathbf{M=m} \rightarrow X=x}$ measures the impact of setting $\mathbf{M}$ to $\mathbf{m}$ along the edges $\boldsymbol{\pi}$ regardless of the value $\mathbf{M}$ is set to along other paths not in $\boldsymbol{\pi}$. Averaging across different values of $\mathbf{m'}$ ensures that the impact is indifferent to the value set along the other paths not in $\boldsymbol{\pi}$. Note that $S^{\mathbb{P}}_{\mathbf{M=m} \rightarrow X=x} \geq 0$ as the multiplicative measure is used in $SE^{\mathbb{P}}_{\boldsymbol{\pi},x}(\mathbf{m, m'})$. Since the scaling factor $S^{\mathbb{P}}_{\mathbf{M=m} \rightarrow X=x}$ measures the impact of $\mathbf{m}$ on $x$ along the direct edges from $M$ to $X$, $S^{\mathbb{P}}_ {\mathbf{M=m} \rightarrow X=x}$ attenuates the influences of paths other than the direct edges from $\mathbf{M}$ to $X$. Using this concept, we formulate edge flow as follows.

\begin{definition}
Edge flow $\mathbb{P}^{\mathbf{M=m}}_{\text{flow}}(X=x)$ is defined as,
\begin{align}
\mathbb{P}^{\mathbf{m}}_{\text{flow}}(x) = \frac{S^{\mathbb{P}}_{\mathbf{m} \rightarrow x} \mathbb{P}(x|do(\mathbf{m}))}{\underset{x}{\sum}S^{\mathbb{P}}_{\mathbf{m} \rightarrow x} \mathbb{P}(x|do(\mathbf{m}))} \label{edgeflow}
\end{align}
\end{definition}
Edge flow $P^{\mathbf{M}}_{\text{flow}}(X)$ is a normalized measure of the scaled probability $S^{\mathbb{P}}_{\mathbf{M} \rightarrow X} \mathbb{P}(X|do(\mathbf{M}))$. Edge flow $P^{\mathbf{M}}_{\text{flow}}(X)$ is a P.M.F \cite{ross2006first} because,
\begin{enumerate}
    \item $0 \leq P^{\mathbf{M}}_{\text{flow}}(X) \leq 1$
    \item $\underset{x}{\sum} P^{\mathbf{M}}_{\text{flow}}(X=x) = 1$
\end{enumerate}

\begin{figure}[h]
\centering
\includegraphics[height=1in]{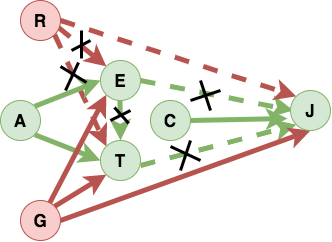}
\caption{Indirect causal influences from $R$ to $J$ in $ \mathbb{P}(J|do(R))$ are attenuated by multiplying with $S^{\mathbb{P}}_{R \rightarrow J}$}. \label{scaling_factor}
\end{figure}

The following conjecture helps in visualizing the dependencies in $S^{\mathbb{P}}_{\mathbf{M} \rightarrow X}\mathbb{P}(X | do(\mathbf{M}))$ along the direct edges $\{M \rightarrow X|M \in \mathbf{M}\}$ in the causal graph $\mathcal{G}$ as shown in Figure \ref{scaling_factor}.
\begin{conjecture}
Let $\mathbb{P}_\mathbf{M}$ be a distribution that factorizes over graph $\mathcal{G}_{\mathbf{M}}$. The active trails $\{M \rightarrow X | M \in \mathbf{M}\}$ are the consequences of the possible dependencies of $X$ and $\mathbf{M}$ in $\mathbb{P}_ {\text{flow}}^{\mathbf{M}}(X | do(\mathbf{M}))$.
\label{conjecture}
\end{conjecture}
The rationale to the above conjecture is as follows. The set of causal paths $\{\text{d-conn}_{\mathcal{G}_{\mathbf{M}}}(X;M|\mathbf{M}\backslash M) | M \in \mathbf{M}\}$ are the active trails that are consequences of the possible dependencies in $\mathbb{P}(X|do(\mathbf{M}))$ i.e. $D^{\mathbb{P}_\mathbf{M}}_{X,\mathbf{M}}=\{(X \not\perp M | \mathbf{M} \backslash M)_{\mathbb{P_\mathbf{M}}} | M \in \mathbf{M}\}$. Since the scaling factor $S^{\mathbb{P}}_{\mathbf{M} \rightarrow X}$ measures the impact on the outcome $X$ when $\mathbf{M}$ is forcibly set to a certain value along the direct paths $\{M \rightarrow X|M \in \mathbf{M}\}$, $S^{\mathbb{P}}_{\mathbf{M} \rightarrow X} \mathbb{P}(X|do(\mathbf{M}))$ attenuates the indirect influences of $X$ on $\mathbf{M}$ in $\mathbb{P}(X|do(\mathbf{M}))$ so as to be influenced only along the direct edges $\{M \rightarrow X|M \in \mathbf{M}\}$. 
$\underset{x}{\sum}S^{\mathbb{P}}_{\mathbf{M} \rightarrow X=x} \mathbb{P}(x|do(\mathbf{M}))$ is only a normalization constant that does not alter the nature of impact of $S^{\mathbb{P}}_{\mathbf{M} \rightarrow X} \mathbb{P}(X|do(\mathbf{M}))$.

The following theorem models the conditional probability distribution $\mathbb{P}(X | Pa(X))$ using a function that inputs the beliefs of $X$ along the fair paths and each of the unfair paths. Note that the beliefs along the unfair paths are separately fed as input to quantify edge unfairness later.

\begin{theorem}
Let $\mathcal{G}$ be a Causal Bayesian Network compatible with $\mathbb{P}*$. Then,
\begin{align}
&\mathbb{P}(X|Pa(X)) = f(P^{\mathbf{F}_X}_{\text{flow}}(X), \underset{A \in \mathbf{U}_{X}} {\bigcup}P^{\mathbf{A}}_{\text{flow}}(X))
\end{align}
s.t,
\begin{align}
&f: \mathbb{W}^{|\mathbf{U}_{X}|+|\mathbf{F}_{X}|+1} \rightarrow [0, 1], \\ &\underset{X}{\sum} f(P^{\mathbf{F}_X}_{\text{flow}}(X), \underset{A \in \mathbf{U}_{X}} {\bigcup}P^{\mathbf{A}}_{\text{flow}}(X)) = 1 \label{eq14}\\
&f(P^{\mathbf{F}_X}_{\text{flow}}(X), \underset{A \in \mathbf{U}_{X}} {\bigcup}P^{\mathbf{A}}_{\text{flow}}(X)) \geq 0
\label{eq15}
\end{align}
where $\mathbf{U}_X$ is the subset of $Pa(X)$ that are along unfair edges and $\mathbf{F}_X$ is the subset of $Pa(X)$ that are along fair edges.
\end{theorem}
\textit{\textbf{Proof:}}
The set of active trails $\mathbf{T}=\{\text{d-conn}_{\mathcal{G}}(X;M|Pa(X)$ $\backslash M)| M \in Pa(X)\}=\{M \rightarrow X | M \in Pa(X)\}$ are the consequences of the possible dependencies $\{(X \not\perp M|Pa(X)\backslash M)|M \in Pa(X)\}$ in $P(X|Pa(X))$. The set $\mathbf{T}_F=\{M \rightarrow X | M \in \mathbf{F}_X\}$ are the consequences of the possible dependencies of $X$ and $\mathbf{F}_X$ in $P^{\mathbf{F}_X}_{\text{flow}}(X)$ [Conjecture 1]. $||_{ly}$, $\mathbf{T}_U=\{A \rightarrow X | A \in \mathbf{U}_X\}$ are the consequences of the possible dependencies of each of $A$ and $X$ in $P^{A}_{\text{flow}}(X)$ $\forall A \in \mathbf{U}_X$ [Conjecture 1]. $\mathbf{T} = \mathbf{T}_F \bigcup \mathbf{T}_U$ or in other words the active trails resulting as a consequence of the dependencies in $P(X|Pa(X))$ are same as the active trails resulting as a consequence of the dependencies in $P^{\mathbf{F}_X}_{\text{flow}}(X)$ and $\underset{A \in \mathbf{U}_{X}}{\bigcup}P^{A}_{\text{flow}}(X)$. Hence, it is reasonable to model $P(X|Pa(X))$ as a function $f$ of $P^{\mathbf{F}_X}_ {\text{flow}}(X)$ and $\underset{A \in \mathbf{U}_{X}} {\bigcup}P^{A}_{\text{flow}}(X)$ provided the function $f$ satisfies the properties of conditional probability $\blacksquare$. 

\subsection{Choices for $f$}
\label{sec:choicesf}
We present two instances for $f$.
\begin{enumerate}[topsep=0pt]
    \item \textbf{Linear Model:} $f$ is a linear combination in the inputs i.e. $\mathbb{P}(X|Pa(X)) \approx f(P^{\mathbf{F}_X}_{\text{flow}}(X), \underset{A \in \mathbf{U}_{X}}{\bigcup}P^{A}_{\text{flow}}(X))=$ 
    \begin{align}
    &\resizebox{.35 \textwidth}{!}{$w_{\mathbf{F}_X \rightarrow X}P^{\mathbf{F}_X}_{\text{flow}}(X) + \underset{A \in \mathbf{U}_{X}}{\sum}w_{A \rightarrow X}P^{A}_{\text{flow}}(X)$} \label{linearmodel}
    \end{align}
    \begin{align}
    s.t., ~&0 \leq w_{\mathbf{F}_X \rightarrow X}, w_{A \rightarrow X} \leq 1, \forall A \in \mathbf{U}_X \label{eq18} \\ &w_{\mathbf{F}_X \rightarrow X} + \underset{A \in \mathbf{U}_{X}}{\sum}w_{A \rightarrow X}=1 \label{eq19}
    \end{align}
    The weight of the belief inputs $w_{\mathbf{F}_X \rightarrow X}$ and $w_{A \rightarrow X}$ are constrained between 0 and 1 since the objective of the mapper $f$ is to capture the interaction between the fraction of the beliefs given by $w_{\mathbf{F}_X \rightarrow X}P^{\mathbf{F}_X}_{\text{flow}}(X)$ and $\underset{A \in \mathbf{U}_{X}}{\bigcup}w_{A \rightarrow X}P^{A}_{\text{flow}}(X)$ to approximate $P(X|Pa(X))$. Eq. (\ref{eq18}) and Eq. (\ref{eq19}) ensure that the conditional probability axioms of $f$ are satisfied. \\ \item \textbf{Neural Network:} $f^{\mathbf{W}}=f^{\mathbf{W}_N}_N \circ ... \circ ~f^{\mathbf{W}_1}_1$ is composite function representing a N-layer neural network with $i^{th}$ layer having $M_i$ neurons and weights $\mathbf{W}_i$ capturing the non-linear combination of the inputs i.e. $\mathbb{P}(X|Pa(X)) \approx$,
    \begin{align}
    &\resizebox{.25 \textwidth}{!}{$f(P^{\mathbf{F}_X}_{\text{flow}}(X), \underset{A \in \mathbf{U}_{X}} {\bigcup}P^{A}_{\text{flow}}(X)) =$} \nonumber \\ 
    &\resizebox{.3 \textwidth}{!}{$f_N(...f_1(P^{\mathbf{F}_X}_{\text{flow}}(X), \underset{A \in \mathbf{U}_{X}}{\bigcup}P^{A}_{\text{flow}}(X)))$} \\
    s.t., ~&0 \leq \mathbf{W}_1 \leq 1, \\ 
    &f_N: \mathbb{R}^{M_N} \rightarrow [0, 1]^{|X|}, \label{eq21}
    \\ &\resizebox{.3 \textwidth}{!}{$\underset{X}{\sum} f(P^{\mathbf{F}_X}_{\text{flow}}(X), \underset{A \in \mathbf{U}_{X}}{\bigcup}P^{A}_{\text{flow}}(X)) = 1$}  \label{eq22}
    \end{align}
    The weights of the belief inputs $\mathbf{W}_1$ are constrained between 0 and 1 since the objective of the mapper $f$ is to capture the interaction between the fraction of the beliefs given by $\mathbf{W}_1[P^{\mathbf{F}_X}_{\text{flow}}(X),\underset{A \in \mathbf{U}_{X}}{\bigcup}P^{A}_{\text{flow}}(X)]$ to approximate $P(X|Pa(X))$. Eq. (\ref{eq21}) and Eq. (\ref{eq22}) ensure that the conditional probability axioms of $f$ are satisfied. One possibility is to use a softmax function for $f_N$ to ensure that the outputs of $f$ satisfy probability axioms. 
\end{enumerate}

\section{Edge Unfairness}
In this section, we use the model formulated in the previous section to quantify edge unfairness and prove that eliminating edge unfairness along all unfair edges results in eliminating cumulative unfairness in any decision towards any subset of sensitive attributes assuming a linear model. 

Edge unfairness $U_{K \rightarrow X}$ is the average unit contribution of edge flow to the approximated $\mathbb{P}(X|Pa(X))$. $U_{K \rightarrow X}$ approximately measures the difference in $\mathbb{P}(X | Pa(X))$ with and without $\mathbb{P}^{K}_{\text{flow}}(X)$ flowing along $K \rightarrow X$.
\begin{definition}
Edge unfairness $U_{e}$ of an unfair edge $e$ is, 
\begin{align}
U_{K \rightarrow X} &=\resizebox{.28 \textwidth}{!}{$\frac{1}{|X,Pa(X)|}\underset{X,Pa(X)}{\sum}\frac{D_{X,K}}{\mathbb{P}^{K}_{\text{flow}}(X)}$} \\ \label{edgeunfair}
\text{where,} ~&D_{X, K}=\resizebox{.28 \textwidth}{!}{$\bigg|f_X(\mathbb{P}^{\mathbf{F}_X}_{\text{flow}}(X), \underset{A \in \mathbf{U}_{X}}{\bigcup}\mathbb{P}^{A}_{\text{flow}}(X))-$} \nonumber \\ &\resizebox{.36 \textwidth}{!}{$f_{X}(\mathbb{P}^{\mathbf{F}_X}_{\text{flow}}(X), \underset{A \in \mathbf{U}_{X}\backslash K}{\bigcup}\mathbb{P}^{A}_{\text{flow}}(X),\mathbb{P}^{K}_{\text{flow}}(X)=0) ~\bigg|$}
\end{align}
\end{definition}
For linear model $f$ (see Section \ref{sec:choicesf}), edge unfairness $U_{K \rightarrow X}$ in edge $K \rightarrow X$ is $w_{K \rightarrow X}$. $w_{K \rightarrow X}$ is a property of the edge and does not vary across different settings of $A$ and $X$.  

To quantify the impact of edge unfairness, we use the notion of discrimination from \cite{zhang2017causal} that measures the cumulative unfairness towards sensitive nodes $\mathbf{S=s}$ in decision $Y=y$. Notations are changed for our usage; direct and indirect discrimination are combined to measure overall unfairness. Cumulative Unfairness $C_{\mathbf{S=s},Y=y}$ is quantified as, 
\begin{align}
    C_{\mathbf{S=s},Y=y} = \frac{1}{|\mathbf{S}\backslash \mathbf{s}|}\underset{\mathbf{s'}\in \mathbf{S}\backslash \mathbf{s}}{\sum} TE_{Y=y}(\mathbf{s, s'})
\end{align}
$|\mathbf{S}\backslash \mathbf{s}|$ is the cardinality of $\mathbf{S}$ without $\mathbf{s}$.
$C_{\mathbf{S=s},Y=y}$ measures the impact on outcome $Y=y$ when $\mathbf{S}$ is forcibly set to $\mathbf{s}$ along the unfair paths from $\mathbf{S}$ to $Y$ irrespective of the value set along other paths. Since all edges emanating from a sensitive node are potential sources of unfairness, total causal effect is used to formulate discrimination. $C_{\mathbf{S=s},Y=y}=0$ shows that the decision $Y=y$ is fair towards $\mathbf{s}$ or $\mathbf{s}$ is treated similar to other groups on average. $C_{\mathbf{S=s},Y=y}\neq0$ shows unfair treatment.

The above theorem points to the fact that eliminating edge unfairness in all unfair edges eliminates cumulative unfairness in any decision towards any subset of sensitive attributes. But, it does not suggest how to remove discrimination.
\begin{theorem}
Cumulative unfairness  $C^{\mathbb{P}}_{\mathbf{S=s},Y=y}$ is zero (no discrimination) when edge unfairness ${U}_{e}=0$, $\forall e$ from $\mathbf{S}$ and when $f(P^{\mathbf{F}_X}_{\text{flow}}(X), \underset{A \in \mathbf{U}_{X}} {\bigcup}P^{A}_{\text{flow}} (X))$ is an error free linear approximation of $\mathbb{P}(X|Pa(X))$ (see Section \ref{sec:choicesf}).
\end{theorem}

\textbf{\textit{Proof:}}\\
\begin{align}
&\mathbb{P}(Y=y~|~do(\mathbf{S=s})) \nonumber\\ &=\underset{\mathbf{V\backslash\{\mathbf{S}},Y\}}{\sum}~~\underset{V \in \mathbf{V} \backslash \{\mathbf{S},Y\}, Y=y}{~\prod}\mathbb{P}(v|pa(V)) \nonumber \\ &[\text{Eq.} ~\ref{factorization} ~\text{a.k.a Factorization formula}] \\ \nonumber \\ &= \underset{\mathbf{V} \backslash (\mathbf{S}, Y)}{\sum} ~\underset{V \in \mathbf{V} \backslash (\mathbf{S},Y), Y=y}{\prod} [w_{\mathbf{F}_V \rightarrow V} P^{\mathbf{F}_V} _{\text{flow}}(V) \nonumber \\ &+ \underset{A \not\in \mathbf{U}_{V} \cap \mathbf{S}}{\sum} w_{A \rightarrow V}P^{A}_{\text{flow}}(V) + \underset{A \in \mathbf{U}_{V} \cap \mathbf{S}}{\sum} w_{A \rightarrow V}P^{\mathbf{s}_{A}}_{\text{flow}}(V)] \nonumber \\ &[f ~\text{is a linear combination in inputs}] \\ \nonumber \\ &= \underset{\mathbf{V} \backslash (\mathbf{S} \bigcup Y)}{\sum} ~\underset{V \in \mathbf{V} \backslash \mathbf{S}, Y=y}{\prod} [w_{\mathbf{F}_V \rightarrow V} P^{\mathbf{F}_V}_{\text{flow}}(V)
\nonumber \\ &+ \underset{A \not\in \mathbf{U}_{V} \cap \mathbf{S}}{\sum} w_{A \rightarrow V}P^{A}_{\text{flow}}(V)] \nonumber \\ &[\text{Since} ~{U}_{e}=0 ~\text{in all unfair edges} ~e ~\text{from} ~\mathbf{S}, {U}_{A \rightarrow V}=\nonumber \\ &w_{A \rightarrow V}=0, ~\forall A \in \mathbf{U}_{V} \cap \mathbf{S}] \label{resproof} \\ \nonumber \\
&\mathbb{P}(Y=y~|~do(\mathbf{S=s}))=\mathbb{P}(Y=y~|~do(\mathbf{S=s'})) \nonumber \\ &[\mathbb{P}(Y=y~|~do(\mathbf{S=s})) ~\text{is independent of} ~\mathbf{s} ~\text{when} ~U_{e}=0 \nonumber \\ ~&\text{in all unfair edges} ~e ~\text{from} ~\mathbf{S} ~\text{as given by Eq.} ~\ref{resproof}] \label{bothequal} \\ \nonumber \\ &TE_{Y=y}(\mathbf{s, s'})=0 \nonumber \\ &[\mathbb{P}(Y=y~|~do(\mathbf{S=s}))=\mathbb{P}(Y=y~|~do(\mathbf{S=s'})) ~\text{from} ~\text{Eq.} \ref{bothequal}] \\ &C^{\mathbb{P}}_{\mathbf{S=s},Y=y} = \frac{1}{|\mathbf{S}\backslash \mathbf{s}|}\underset{\mathbf{s'}\in\mathbf{S}\backslash\mathbf{s}}{\sum} TE_{Y=y}(\mathbf{s, s'})=0 \nonumber \\ &[\text{No discrimination}] \blacksquare
\end{align}

We now measure the potential to mitigate cumulative unfairness when edge unfairness is decreased. Using the measure of potential and the amount of edge unfairness, we formulate a priority algorithm. 

The following quantity measures the variation in $C^{\mathbb{P},\text{approx}}_{\mathbf{S=s},Y=y}$ as edge unfairness $U_{e}$ in edge $e$ is varied. $C^{\mathbb{P},\text{approx}} _{\mathbf{S=s},Y=y}$ is the cumulative unfairness when $CPTs$ are replaced by their functional approximation $f(P^{\mathbf{F}_X} _{\text{flow}}(X), \underset{A \in \mathbf{U}_{X}} {\bigcup}P^{A}_ {\text{flow}}(X))$ under least squares. 
\begin{definition}
\textbf{Sensitivity} of $C^{\mathbb{P},\text{approx}}_{\mathbf{S=s},Y=y}$ w.r.t $U_{e}$ is,
\begin{align}
&S^{\mathbf{S=s},Y=y}_{e}=\frac{\partial C^{\mathbb{P},\text{approx}}_{\mathbf{S=s},Y=y}}{\partial \mu_{e}} \bigg|_{\mathbf{U} _{\text{current}}} \label{sens}
\end{align} 
where $\boldsymbol{U}_{\text{current}}$ is the edge unfairness obtained by decomposing the observational distribution $\mathbb{P}(\mathbf{V})$.
\end{definition}
The following quantity measures the potential to mitigate $C^{\mathbb{P},\text{approx}} _{\mathbf{S=s},Y=y}$ when edge unfairness $U_{e}$ in edge $e$ is decreased.
\begin{definition}
\textbf{Potential} of $C^{\mathbb{P},\text{approx}} _{\mathbf{S=s},Y=y}$ to move towards 0 (indicative of non-discrimination) when $U_{e}$ is decreased is,
\[
    P^{\mathbf{S=s},Y=y}_{e} =
    \begin{cases}
        \bigg|S^{\mathbf{S=s},Y=y}_{e}\bigg| & \text{if}~C^{\mathbb{P},\text{approx}}_{\mathbf{S=s},Y=y}=0\\ \\
        S^{\mathbf{S=s},Y=y}_{e} & \text{if}~ C^{\mathbb{P},\text{approx}}_{\mathbf{S=s},Y=y} > 0\\ \\
        -S^{\mathbf{S=s},Y=y}_{e} & \text{if}~ C^{\mathbb{P},\text{approx}}_{\mathbf{S=s},Y=y} < 0
    \end{cases} \label{sens}
\]
\end{definition}
The intuition of $P^{\mathbf{S=s},Y=y}_{e}$ is that if $C^{\mathbb{P},\text{approx}}_{\mathbf{S=s},Y=y}=0$, then $C^{\mathbb{P},\text{approx}}_{\mathbf{S=s},Y=y}$ moves away from 0 when edge unfairness in edge $e$ is decreased either by getting reduced or increased as quantified by $|S^{\mathbf{S=s},Y=y}_{e}|$. Other scenarios can be analyzed similarly. Higher order derivatives of $C^{\mathbb{P},\text{approx}}_{\mathbf{S=s},Y=y}$ with respect to edge unfairness is 0. 
\section{Unfair Edges Prioritization and Discrimination Removal} \label{algorithms}
Based on the aforementioned theorems and the definitions in the previous sections, we present the following pseudo-codes for computing the edge unfairness, prioritize the unfair edges, and remove discrimination from the causal bayesian network $(\mathbb{G}, \mathbb{P})$.

\begin{algorithm}[!h]
    \caption{fitCPT($\mathbb{G}$,$\mathbb{P}$,$\mathbf{E}^{\text{unfair}}_{\mathbb{G}}$,X)}
    \begin{algorithmic}[1]
    \State \text{Initialize} $\mathbf{w}$ \text{randomly}
    \State $\mathbf{Y}$ $\leftarrow \mathbb{P}(X=x|Pa(X)=pa(X))$
    \State $\text{Compute}~\mathbb{P}^{\mathbf{F}_X}_{\text{flow}}(X)$ \text{and} $\mathbb{P}^{A}_{\text{flow}}(X) ~\forall A \in \mathbf{U}_X \hfill \text{(Def. \ref{edgeflow})}$
    \State $\hat{\mathbf{Y}}(\mathbf{w}) \leftarrow f^\mathbf{w}(\mathbb{P}^{\mathbf{F}_X}_{\text{flow}}(X), \underset{A \in \mathbf{U}_X}{\bigcup}\mathbb{P}^{A}_{\text{flow}}(X))$
    \State $\mathbf{w}^* \leftarrow \text{arg}\min_{\mathbf{w}}||\mathbf{Y} - \hat{\mathbf{Y}}(\mathbf{w})||^2$ \text{subject to} Eq.\ref{eq18} \& \ref{eq19}\\
    \textbf{Output:} $\mathbf{w}^*$
    \end{algorithmic} \label{fitCPT}
\end{algorithm}

\begin{algorithm}[!h]
    \caption{unfairEdgePriority($\mathbb{G},\mathbb{P},\mathbf{E}^{\text{unfair}}_{\mathbb{G}}$,$\mathbf{s}$,$y$,$w_u$,$w_{p}$)}
    \begin{algorithmic}[1]
    \State $\mathbf{w}^{*} = \{\}$ \Comment{Weights of the approximated $CPTs$}
    \For{$V~ \text{in}~ \mathbf{V}$}
        \State $\mathbf{w}^*_V \leftarrow \text{fitCPT}(\mathbb{G},\mathbb{P},\mathbf{E}^{\text{unfair}}_{\mathbb{G}},V)$ \Comment{See Algorithm \ref{fitCPT}}
        \State $\mathbf{w}^{*} \leftarrow \mathbf{w}^{*}\cup \{\mathbf{w}^*_V\}$
    \EndFor
    \State priorityList = $\{\}$
    \For{$e=S \rightarrow V ~\text{in} ~\mathbf{E}^{\text{unfair}}_{\mathbb{G}}$}
      \State Compute $U_{e}$ using $f^{\mathbf{w}^{*}_{V}}$ from Eq. \ref{edgeunfair}
      \State Compute $P^{\textbf{S=s},Y=y}_{e}$ \hfill (Def. \ref{sens})
    \EndFor
    \State \text{priority} = $w_uU_{e} + w_{p} P^{\textbf{S=s},Y=y}_{e}$
    \State \text{priorityList = priorityList} $\cup \{(e,\text{priority)}\}$ \\
    \textbf{Output:} $\text{priorityList}$
    \end{algorithmic} \label{priority}
\end{algorithm}
    
\begin{algorithm}[!h]
\caption{removeDiscrimination($\mathbb{G}$,$\mathbb{P}$,$\mathbf{E}^{\text{unfair}}_{\mathbb{G}}$)}
\begin{algorithmic}[1]
\State \resizebox{.4 \textwidth}{!}{$\tilde{\mathbf{w}} \leftarrow \text{argmin}_{\mathbf{w}} \underset{e \in \mathbf{E}^{\text{unfair}}_{\mathbb{G}}}{\mathlarger{\sum}} U_{e, \mathbb{G}} + \left\lVert \mathbb{P}(\mathbf{V}) - \underset{Z \in \mathbf{V}}{\mathlarger{\prod}} \hat{\mathbf{Y}}_Z({\mathbf{w}})\right\rVert^2$} \hspace*{7mm}~\text{subject to Eq.} ~\ref{eq18} ~\text{and Eq.} ~\ref{eq19}
\State \resizebox{.18 \textwidth}{!} {$\mathbb{P}_{\text{new}}(\mathbf{V}) \leftarrow \underset{X\in \mathbf{V}} {\mathlarger{\prod}} \hat{\mathbf{Y}}_X({\tilde{\mathbf{w}}})$}\\
\textbf{Output:} $\mathbb{P}_{\text{new}}(\mathbf{V})$
\end{algorithmic} \label{disc_algo}
\end{algorithm}
\begin{enumerate}
    \item \textbf{fitCPT() Algorithm \ref{fitCPT}}: It takes the causal model $(\mathbb{G},\mathbb{P})$, the set of unfair edges $\mathbf{E}^{\text{unfair}}_{\mathbb{G}}$ and the attribute $X$ as inputs and approximates the $\mathbb{P}(X|Pa(X))$ by the model $f^\mathbf{w}$ using the least squares loss. 
    
    \item \textbf{unfairEdgePriority() Algorithm \ref{priority}}: It computes the priorities of the unfair edges based on their edge unfairness and potential to mitigate the cumulative unfairness on reducing-edge unfairness. This cumulative unfairness can then be used by law enforcement to address underlying unfairness.
    
    \item \textbf{removeDiscrimination() Algorithm \ref{disc_algo}}: It removes discrimination by regenerating new $CPTs$ for a causal model $(\mathbb{G},\mathbb{P})$ with given set of unfair edges $\mathbf{E}^{\text{unfair}}_{\mathbb{G}}$. These $CPTs$ are approximated by solving an optimization problem of minimizing the overall edge unfairness subject to the constraints that the axioms of the conditional probabilities are satisfied. A \textit{data utility} term, that is essentially the $MSE$ between $\mathbb{P}(\mathbf{V})$ and the new joint distribution computed by the product of approximated $CPTs$, is added to the objective function to ensure that the influences from other insensitive nodes are preserved.
\end{enumerate}

\section{EXPERIMENTS}
\label{section:exper}
In this section, we describe the experiments that were performed to validate the model assumption. Consider the causal graph for the criminal recidivism problem in Figure \ref{main_graph}. In this section below, we describe this causal model.
\subsection{Causal model}
We construct the causal graph along the lines of the structural relationships given by the authors in \cite{vanderweele2011causal}. The difference is that the graph described in their paper contains judge attributes while our work contains accused attributes. That is, we consider judicial decisions made based on case characteristics and attributes of the accused such as criminal's race, gender, age, etc. In such a scenario, unfairness arises when the bail decision is taken based on the sensitive attributes such as the race and gender of the accused. The values taken by the variables are,
\begin{table}[h!]
    \centering
    \caption{Nodes and their corresponding values.}\label{att_table}
    \resizebox{\linewidth}{!}{\begin{tabular}{c|c}
    \textbf{Node} & \textbf{Values}\\
        \hline
        Race $R$ & African American($0$), Hispanic($1$) and White($2$) \\
        Gender $G$ & Male($0$), Female($1$) and Others($2$) \\
        Age $A$ & Old ($0$)($>$35y) and Young ($1$) ($\leq$ 35y) \\
        Education $E$ & Literate ($0$) and Illiterate ($1$) \\
        Training $E$ & Not Employed ($0$) and Employed ($1$) \\
        Bail Decision $J$ & Bail granted ($0$) and Bail rejected ($1$) \\
        Case History $C$ & Strong ($0$) and Weak criminal history ($1$)
    \end{tabular}}
\end{table}

Conditional probability distribution or table ($CPT$) of an attribute $V$ is $\mathbb{P}(V|Pa(V))$. We now describe how $CPT$ is generated. For this, we define the following quantities,
\begin{enumerate}
    \item \textit{Parameters}: $\theta_{A\rightarrow V} \in [0, 1] ~\forall V \in ~\mathbf{V}, \forall A \in Pa(V)$ where $\theta_{A \rightarrow V}$ quantifies the direct influence of attribute $A$ on $V$ that is not dependent on the specific values taken by $A$ and $V$. $\theta_{A \rightarrow V}$ is an edge property.
    \item \textit{Scores}: $\lambda_{A=a\rightarrow V=v} \in [0,1] ~\forall V \in ~\mathbf{V}, \forall A \in Pa(V)$ where $\lambda_{A=a\rightarrow V=v}$ quantifies the direct influence of attribute $A$ on $V$. It is sensitive to $A$ and $V$ values.
\end{enumerate}
We generate the CPT of a variable $V$ by computing the weighted sum of all the scores $\lambda_{A=a\rightarrow V=v}$ where the weights are the parameters $\theta_{A\rightarrow V}$. That is,
\begin{align}
\mathbb{P}(v | pa(V)) = \sum_{A \in Pa(V)}\theta_{A\rightarrow V}\lambda_{A=a\rightarrow V=v} \label{modeleq}
\end{align}

For instance, taking an example of Training ($T$),
\begin{align}
\mathbb{P}(T=t| R=r, A=a, G=g) =
\theta_{R\rightarrow T} \lambda_{R=r\rightarrow T=t} \nonumber \\+\theta_{A\rightarrow T} \lambda_{A=a\rightarrow T=t} + \theta_{G\rightarrow T} \lambda_{G=g\rightarrow T=t}
\end{align}
To ensure that the generated $CPTs$ satisfy the marginality condition, we define the following constraints over the parameters and scores,
\begin{align}
\sum_{A\in Pa(T)} \theta_{A\rightarrow T} &= 1 \\
\sum_{t} \lambda_{A=a\rightarrow T=t} &= 1,~~\forall A\in Pa(T)
\end{align}

\subsection{Measuring Edge Unfairness}
Once the $CPTs$ are constructed for all $V\in\mathbf{V}$, we fit Eq. \ref{linearmodel} for every CPT by solving the constrained least-squares problem (CLSP) to find the optimal solution. We solve CLSP to obtain $\mathbf{w}^*$ by following the steps given in Algorithm \ref{priority}. The CLSP is a well-known optimization problem whose implementation is available in \textsc{scikit-learn} library in Python. 

\textbf{Inference:} Now that the parameterized model is defined, we repeat the same algorithm for $625$ distinct combinations of $\{\theta_{A\rightarrow J}, \theta_{B \rightarrow T} | A \in Pa(J), B \in Pa(T)\}$. In all cases, we found that the optimal weights $\mathbf{w}^*$ a.k.a the edge unfairness obtained are approximately equal to the $\{\theta_{P_J\rightarrow J}, \theta_{P_T \rightarrow T} | P_J \in Pa(J), P_T \in Pa(T)\}$. For example, when we set $\{\theta_{P_J\rightarrow J} | P_J \in Pa(J)\} = \{\frac{1}{|Pa(J)|}\}^{|Pa(J)|}$ i.e., equal strength to all parents and solve the CLSP, we obtained $\mathbf{w}^* \approx \{0.3, 0.4, 0.3\}$ with a mean squared error ($e_J$) of order $10^{-3}$. A low MSE indicates that the edge unfairness is quantifiable using a linear model. Non-linear model would be required to capture the edge unfairness if the $CPTs$ we generated by a non-linear combination inputs.

\subsection{Edge Unfairness is insensitive to Attribute Values}
Edge unfairness is a property of the edge and its value should not depend on the values taken by the attributes corresponding to it. To validate it, we compare the edge unfairness $w^*_{R\rightarrow J}$ and $w^*_{G\rightarrow J}$ obtained by solving CLSPs with the $\theta_{R\rightarrow J}$ and $\theta_{G\rightarrow J}$ respectively. 

\begin{figure}[h!]
    \centering
    \includegraphics[width=\linewidth]{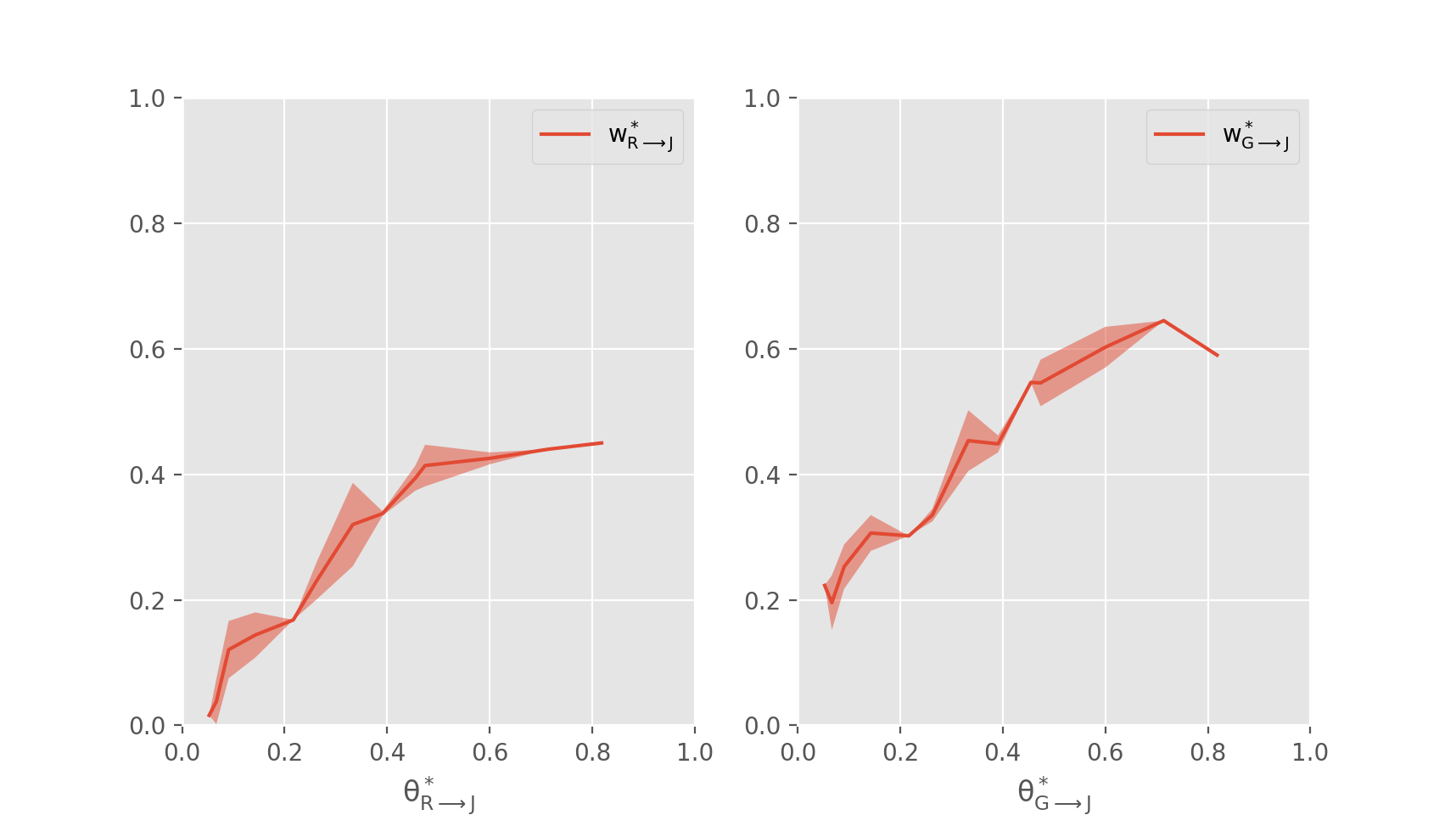} \caption{$w^*_{R\rightarrow J}$ vs. $\theta_{R\rightarrow J}$ and $w^*_{G\rightarrow J}$ vs. $\theta_{G\rightarrow J}$}
    \label{theta_weight}
\end{figure}

\textbf{Inference:} From the Figure \ref{theta_weight} below, one can infer that the edge unfairness $\mathbf{w}^*$ is insensitive to the specific values taken by the parent attributes as expected. For instance, out of $625$ combinations of $\{\theta_{A\rightarrow J}, \theta_{B \rightarrow T} | A \in Pa(J), B \in Pa(T)\}$, the optimal weights $w^*_{R\rightarrow J}$ obtained in all the models with $\theta_{R\rightarrow J} = 0.33$ are in the range $[0.33, 0.40]$. A small deviation in $w^*_{R\rightarrow J}$ shows that $w^*_{R\rightarrow J}$ depends only on $\theta_{R\rightarrow J}$ and not on the specific values taken by the attributes.

\subsection{Impact by Scaling Factor}
\begin{figure*}[htp]
  \subfigure{\includegraphics[width=0.33\textwidth]
  {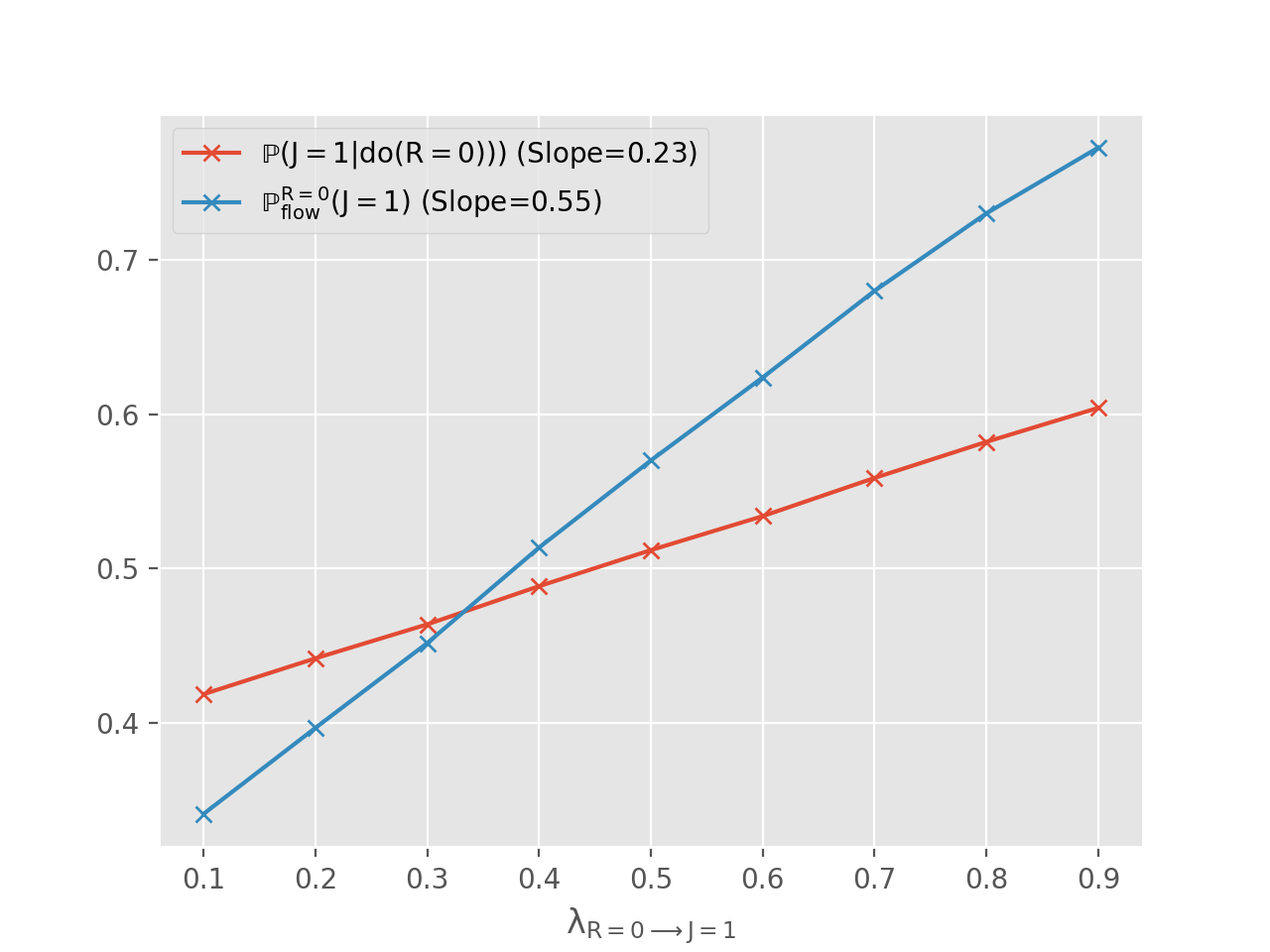}}
  \quad
  \subfigure{\includegraphics[width=0.33\textwidth]
  {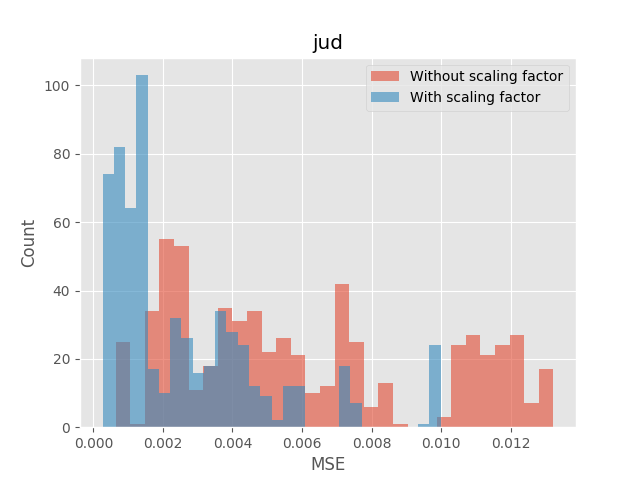}}
  \quad
  \subfigure{\includegraphics[width=0.33\textwidth]
  {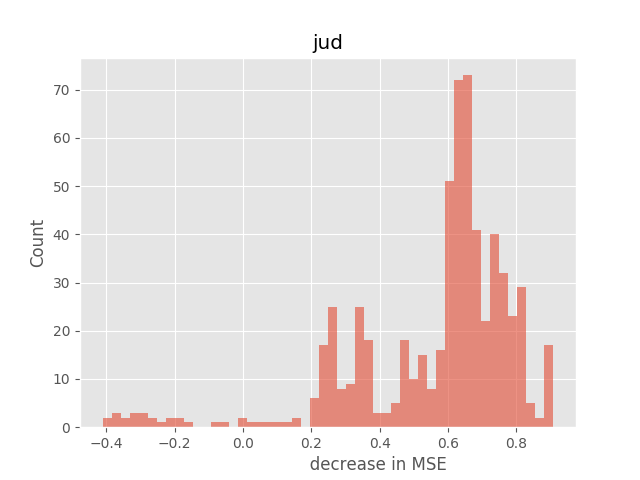}} 
  \caption{(a)Inputs to the model with Scaling factor (blue curve) and without the Scaling factor (red curve) by varying $\lambda_{R=0\rightarrow J=1}$. (b)Histogram for without scaling factor $e'_J$ (red) and with scaling factor $e_J$ (blue). (c)Histogram for $\delta_J$ (\% decrease in MSE).}
  \label{mse_decrease}
\end{figure*}

In the previous section it was established that the values of edge unfairness $\mathbf{w}^*$ are close to the $\{\theta_{e} | e \in \mathbf{E}^{\text{unfair}}_{\mathcal{G}}\}$. This section discusses the effect of introducing Scaling factors into the model input. We first show that the inputs to the model $f$ are correlated with the Score ($\lambda_{A=a\rightarrow V=v}$) and the Scaling factor increases the correlation even more. For this experiment, $\mathbb{P}(J=1|pa(J))$ is decomposed using CLSP. Here, $\lambda_{R=0\rightarrow J=1}$ is varied and and the other quantities used to generate $\mathbb{P}(J=1|pa(J))$ are fixed to certain value. For a given $\lambda_{R=0\rightarrow J=1}$, the following two quantities are computed,
\begin{enumerate}
    \item $\mathbb{P} (J=1|do(R=0))$
    \item $\mathbb{P}^{R=0}_{\text{flow}}(J=1)$
\end{enumerate}
Figure \ref{mse_decrease} plots the above quantities alongside $\lambda_{R=0\rightarrow J=1}$. 

\textbf{Inference:}
We observe that both the quantities are linearly dependent on $\lambda_{R=0\rightarrow J=1}$. But the latter $\mathbb{P}^{R=0}_{\text{flow}}(J=1)$ has a slope=0.55 as compared to the former $\mathbb{P} (J=1|do(R=0))$ that has a slope=0.23 $<< 0.5$ and is therefore a better representation of $\lambda_{R=0\rightarrow J=1}$. This validates that using Scaling factors in the inputs leads to  better approximation of the inputs to $\lambda_{A=a\rightarrow V=v}$.

Next, to validate that the scaling factor improves the model performance, CLSP was solved for 625 combinations of $\{\theta_{A\rightarrow J}, \theta_{B \rightarrow T} | A \in Pa(J), B \in Pa(T)\}$ and the Mean Squared Errors (MSEs) between the CPT for Judicial bail $\mathbb{P}(J|R, G, C, E, T)$ and its linear functional approximation $f$ (see Eq. \ref{linearmodel}) were recorded. This experiment was performed for the following settings,
\begin{enumerate}
\item MSEs calculated by feeding the inputs to the model without the use of scaling factor denoted by $e'_J$ \item MSEs calculated by feeding the inputs to the model by using the scaling factor denoted by $e_J$
\end{enumerate}
\textbf{Inference:} 
Distributions of $e'_J$ and $e_J$ are plotted in  Figure \ref{mse_decrease}. Here, the maximum value of $e'_J$ (red bar) is obtained around $0.013$ and it is \textit{evenly} distributed in the range $(0.0,0.014)$. On the other hand, $e_J$ (blue bars) is \textit{skewed} in the lower error range i.e., $(0.0,0.004)$ with the maximum value of $e_J$ (blue bar) obtained around $0.01$. 
This means that the usage of scaling factors in Def. \label{def_linear} is a better choice because the MSEs distribution is skewed in the lower error range of $(0.0,0.004)$ with the scaling factor as compared to the other case. To analyze the extent to which MSE is reduced after introducing the Scaling factor, we calculate the percentage decrease in the MSEs $\delta_J$ equal to, 
\begin{align}
\delta_J= \frac{e'_J-e_J}{e'_J}
\end{align}
and plot its distribution in the Figure \ref{mse_decrease}. As seen from Figure \ref{mse_decrease}, the majority of the values of $\ delta_J$ are around $60\%-70\%$ which is a significant decrease in terms of the MSEs.

Another observation from this experiment is that $\delta_J$ is negative in few settings. Those specific settings are the instances where $\theta_{A\rightarrow  V}$ ($V\in \{J, T\}$) is large for a particular $A \in Pa(V)$ and other $\theta_{A'\rightarrow  V}$ where $A'\in Pa(V)\setminus A$ are negligible. Example of such a case is $\{\theta_{A\rightarrow J} | A \in Pa(J)\} = \{\theta_{R\rightarrow J}, \theta_{G\rightarrow J}, \theta_{C\rightarrow J}, \theta_{E\rightarrow J}, \theta_{T\rightarrow J}\} = \{0.81, 0.09, 0.03, 0.03, 0.03\}$. $\delta_J$ is negative only $4\%$ of the $625$ combinations of $\boldsymbol{\theta}= \{\theta_{A\rightarrow J}, \theta_{B \rightarrow T} | A \in Pa(J), B \in Pa(T)\}$. This shows that the usage of scaling factor in the model input decreases the MSEs in almost all combinations of $\boldsymbol{\theta}$. Moreover, all those $4\%$ negative $\delta_J$ are observed in extreme cases as illustrated in the above example. One example of such a case could be the judicial decision made entirely based on a sensitive attribute, say Race of the criminal, without considering any other factors. But such situations rarely occur in practice.

\subsection{Finite data}
\begin{figure}[H]
    \centering
    \includegraphics[width=0.55\linewidth]{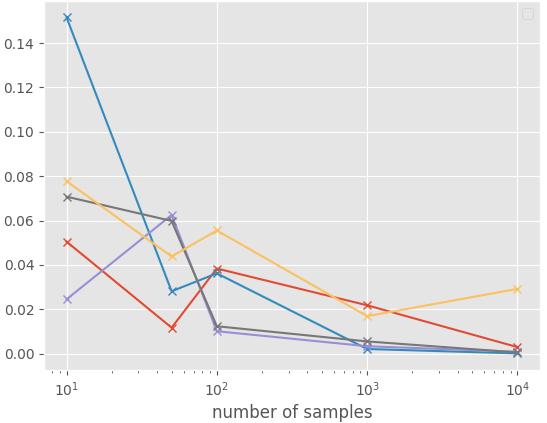}
    \caption{$||\mathbf{w}^*(\mathbb{P}) - \mathbf{w}^*(\mathbb{P}^m)||_2$ vs number of samples $m$}
    \label{finite}
\end{figure}

In this section, we examine the applicability of the proposed approach to realistic scenarios.
In most of the real-world settings knowledge of the causal model and the $CPTs$ are not available. Since our work assumes that the causal graph is given, we do dwell on discovering causal structures using a finite amount of data. However, we evaluate the impact of a finite amount of data on $CPTs$ and their impact on calculating edge unfairness. Edge unfairness $\mathbf{w}^*(\mathbb{P})$ calculated using Algorithm \ref{priority} using true $CPTs$ $\mathbb{P}$ is compared with the edge unfairness $\mathbf{w}^*(\mathbb{P}^m)$ calculated using Algorithm \ref{priority} using estimated $CPTs$ $\mathbb{P}^m$ where $m$ is the number of samples used for estimation by maximizing the likelihood. $m$ samples are drawn from $\mathbb{P}$ randomly.
For a given $m$, the distance between $\mathbf{w}^*(\mathbb{P})$ and $\mathbf{w}^*(\mathbb{P}^m)$ is calculated using Euclidean distance $E(\mathbb{P},\mathbb{P}^m) = ||\mathbf{w}^*(\mathbb{P}) - \mathbf{w}^*(\mathbb{P}^m)||_2$. We repeat this experiment for different $m$ and compute $E(\mathbb{P},\mathbb{P}^m)$. Intuitively, the distance should decrease as $m$ increases, because a large number of i.i.d. samples produces a better approximation of the original distribution $\mathbb{P}$, thereby reducing the distance between $\mathbf{w}^*(\mathbb{P}^m)$ and $\mathbf{w}^*(\mathbb{P})$. $E(\mathbb{P},\mathbb{P}^m)$ is plotted against $m$ in Figure \ref{finite} for different true distributions $\mathbb{P}$ that are randomly generated (different colours). 
\textbf{Inference:} We observe that $\mathbf{w}^*(\mathbb{P}^m)$ move closer to $\mathbf{w}^*(\mathbb{P})$ as $m$ increases. Moreover, since $\mathbb{P}$ was randomly generated, we also observe that there exists an empirical bound over $E(\mathbb{P}, \mathbb{P}^m)$ for a given $m$. For instance, in the Figure \ref{finite}, for $m$ greater than $10^3$, the observed $E(\mathbb{P},\mathbb{P}^m)$ is always less than $0.04$ and this bound shows a decreasing trend as $m$ increases for any true $\mathbb{P}$. Hence, given the number of samples $m$, we can get an empirical upper bound over the error between the true and the estimated edge unfairness which helps in evaluating whether the calculated edge unfairness should be used in real-life scenarios where only a finite amount of data is available. The presence of empirical bound motivates the reader to investigate the possibility of a theoretical bound over $E(\mathbb{P}, \mathbb{P}^m)$. Theoretical bound is not presented in our work. 
\section{Related Work}
\textbf{Mitigating Unfairness in the Data Generation Phase:} \cite{gebru2018datasheets} suggests documenting the dataset by recording the motivation and creation procedure. It does not quantify the contribution of potential sources of unfairness towards discrimination.\\
\textbf{Assumptions:}  We assume that the causal graph contains only observed variables unlike \cite{tian2003studies}. \cite{zhang2017causal} assumes that the sensitive variable $S$ has no parents as it is an inherent nature of the individual. We follow \cite{zhang2019boundscausal} which relaxes this assumption because sensitive nodes such as religious belief can have parents like literacy $L$. Our paper assumes a discrete node setting to avoid deviating from our primary objective. \cite{nabi2018fair} discuss a discrimination removal procedure in a continuous node setting by altering data distribution. They discuss different ways to handle non-identifiability and then determine the outcome of future samples that come from an unfair distribution. Nevertheless, they do not discuss how to mitigate unfairness underlying potential sources of unfairness.\\
\textbf{Edge Flow:} Decomposing direct parental dependencies on a child into independent contributions from each of the parent's aids in quantifying the edge flow. \cite{srinivas1993generalization, kim1983computational, henrion2013practical} separate the independent contributions from each of the parents onto the child by using unobserved nodes in the representation of causal independence. To overcome the issues of intractability and unobserved nodes, \cite{heckerman1993causal} proposed a temporal definition of causal independence which states that if only the cause $c$ transitions from time $t$ to $t+1$, then the effect's distribution at time $t+1$ depends only on the effect and the cause at time $t$, and the cause at time $t+1$. Based on this definition, a belief network representation is constructed with the observed nodes that make the probability assessment and inference tractable. \cite{heckerman1994new} proposes a temporal equivalent to the temporal definition of \cite{heckerman1993causal}. The aforementioned works do not decompose direct dependencies from each of the parents onto the child as in our work.\\
\textbf{Edge Unfairness:} Multiple statistical criteria have been proposed to identify discrimination (\cite{berk2018fairness}) but it is mathematically incompatible to satisfy them all. Consequently, there is an additional task of selecting which criterion has to be achieved. Moreover, statistical criteria do not help in identifying the sources of unfairness. \cite{zhang2017causal} use path-specific effects to identify direct and indirect discrimination after data is generated but miss on how to mitigate unfairness in the data generation phase. They assume that the indirect paths contain a redlining node and direct paths contain a sensitive node as the source is unfair. Still, they do not pinpoint the potential sources of unfairness that are of interest in our work. For instance, all paths from gender $G$ to bail decision $J$ via literacy $L$ are unfair as $L$ is a redlining node. But the potential sources of unfairness are edges emanating from $G$ each varying in the type of unfairness. 
\cite{chiappa2018causal} use the notion of an unfair edge to make the sources of unfairness explicit. We combine the direct and indirect paths into unfair paths since our objective is to quantify the impact of potential sources of unfairness on a decision that occurs via all the unfair paths.\\
\textbf{Discrimination Removal Procedure:} \cite{zhang2017causal} removes discrimination by altering the data distribution through an optimizing procedure with anti-discrimination constraints. \textit{Firstly}, these non-linear constraints grow exponentially in the number of nodes (and values taken by the sensitive nodes) that eventually increases the time to solve the quadratic programming problem. 
\textit{Secondly}, these constraints depend on a subjectively chosen threshold of discrimination that is disadvantageous because the regenerated data distribution would remain unfair had a smaller threshold been chosen. Our paper formulates a discrimination removal procedure without exponentially growing constraints and a threshold of discrimination.

\section{CONCLUSION}
We introduce the problem of quantifying belief flowing along an unfair edge and show the impact of edge unfairness on the cumulative unfairness in a causal graph. The theorem motivates law enforcement to mitigate cumulative unfairness by reducing the edge unfairness. We present an unfair edge priority computation algorithm that can be used by law enforcement. We also design a discrimination removal algorithm by eliminating the constraints that grow exponentially in the number of sensitive nodes and the values taken by them. Our work gives tangible directions for law enforcement to mitigate unfairness underlying the unfair edges. There is no utility in making cautionary claims of discrimination when it is not complemented with information that points to the potential sources of unfairness causing discrimination. This paper attempts to provide such concrete information. In the future, we aim to extend our work to the non-parametric setting, evaluate the impact of edge unfairness on subsequent stages of the machine learning pipeline such as selection, classification, etc. We also plan to extend to the semi-Markovian \cite{wu2019pc} and continuous nodes settings \cite{nabi2018fair}.
\subsubsection*{Acknowledgements}
This work was funded by the Robert Bosch Centre for Data Science and Artificial Intelligence Lab at IIT Madras. We would like to thank Prof. Harish Guruprasad Ramaswamy for the valuable feedback that greatly assisted this research. 

\bibliography{maintext} 

\begin{thebibliography}{}

\bibitem[Act, 1964]{act1964civil}
Act, C.~R. (1964).
\newblock Civil rights act of 1964.
\newblock {\em Title VII, Equal Employment Opportunities}.

\bibitem[Avin et~al., 2005]{avin2005identifiability}
Avin, C., Shpitser, I., and Pearl, J. (2005).
\newblock Identifiability of path-specific effects.

\bibitem[Barocas and Selbst, 2016]{barocas2016big}
Barocas, S. and Selbst, A.~D. (2016).
\newblock Big data's disparate impact.
\newblock {\em Calif. L. Rev.}, 104:671.

\bibitem[Berk et~al., 2018]{berk2018fairness}
Berk, R., Heidari, H., Jabbari, S., Kearns, M., and Roth, A. (2018).
\newblock Fairness in criminal justice risk assessments: The state of the art.
\newblock {\em Sociological Methods \& Research}, page 0049124118782533.

\bibitem[Chiappa and Isaac, 2018]{chiappa2018causal}
Chiappa, S. and Isaac, W.~S. (2018).
\newblock A causal bayesian networks viewpoint on fairness.
\newblock In {\em IFIP International Summer School on Privacy and Identity
  Management}, pages 3--20. Springer.

\bibitem[Gebru et~al., 2018]{gebru2018datasheets}
Gebru, T., Morgenstern, J., Vecchione, B., Vaughan, J.~W., Wallach, H.,
  Daume{\'e}~III, H., and Crawford, K. (2018).
\newblock Datasheets for datasets.
\newblock {\em arXiv preprint arXiv:1803.09010}.

\bibitem[Heckerman, 1993]{heckerman1993causal}
Heckerman, D. (1993).
\newblock Causal independence for knowledge acquisition and inference.
\newblock In {\em Uncertainty in Artificial Intelligence}, pages 122--127.
  Elsevier.

\bibitem[Heckerman and Breese, 1994]{heckerman1994new}
Heckerman, D. and Breese, J.~S. (1994).
\newblock A new look at causal independence.
\newblock In {\em Uncertainty Proceedings 1994}, pages 286--292. Elsevier.

\bibitem[Henrion, 2013]{henrion2013practical}
Henrion, M. (2013).
\newblock Practical issues in constructing a bayes' belief network.
\newblock {\em arXiv preprint arXiv:1304.2725}.

\bibitem[Kim and Pearl, 1983]{kim1983computational}
Kim, J. and Pearl, J. (1983).
\newblock A computational model for causal and diagnostic reasoning in
  inference systems.
\newblock In {\em International Joint Conference on Artificial Intelligence},
  pages 0--0.

\bibitem[Koller and Friedman, 2009]{koller2009probabilistic}
Koller, D. and Friedman, N. (2009).
\newblock {\em Probabilistic graphical models: principles and techniques}.
\newblock MIT press.

\bibitem[Maathuis et~al., 2018]{maathuis2018handbook}
Maathuis, M., Drton, M., Lauritzen, S., and Wainwright, M. (2018).
\newblock {\em Handbook of graphical models}.
\newblock CRC Press.

\bibitem[Nabi and Shpitser, 2018]{nabi2018fair}
Nabi, R. and Shpitser, I. (2018).
\newblock Fair inference on outcomes.
\newblock In {\em Thirty-Second AAAI Conference on Artificial Intelligence}.

\bibitem[Pearl, 2009]{pearl2009causality}
Pearl, J. (2009).
\newblock {\em Causality}.
\newblock Cambridge university press.

\bibitem[Ross et~al., 2006]{ross2006first}
Ross, S.~M. et~al. (2006).
\newblock {\em A first course in probability}, volume~7.
\newblock Pearson Prentice Hall Upper Saddle River, NJ.

\bibitem[Shpitser, 2013]{shpitser2013counterfactual}
Shpitser, I. (2013).
\newblock Counterfactual graphical models for longitudinal mediation analysis
  with unobserved confounding.
\newblock {\em Cognitive science}, 37(6):1011--1035.

\bibitem[Srinivas, 1993]{srinivas1993generalization}
Srinivas, S. (1993).
\newblock A generalization of the noisy-or model.
\newblock In {\em Uncertainty in artificial intelligence}, pages 208--215.
  Elsevier.

\bibitem[Tian, 2003]{tian2003studies}
Tian, J. (2003).
\newblock Studies in causal reasoning and learning.

\bibitem[VanderWeele and Staudt, 2011]{vanderweele2011causal}
VanderWeele, T.~J. and Staudt, N. (2011).
\newblock Causal diagrams for empirical legal research: a methodology for
  identifying causation, avoiding bias and interpreting results.
\newblock {\em Law, Probability \& Risk}, 10(4):329--354.

\bibitem[Wu et~al., 2019]{wu2019pc}
Wu, Y., Zhang, L., Wu, X., and Tong, H. (2019).
\newblock Pc-fairness: A unified framework for measuring causality-based
  fairness.
\newblock In {\em Advances in Neural Information Processing Systems}, pages
  3404--3414.

\bibitem[Zhang et~al., 2017]{zhang2017causal}
Zhang, L., Wu, Y., and Wu, X. (2017).
\newblock A causal framework for discovering and removing direct and indirect
  discrimination.
\newblock In {\em Proceedings of the Twenty-Sixth International Joint
  Conference on Artificial Intelligence}.

\bibitem[{Zhang} et~al., 2019]{zhang2019boundscausal}
{Zhang}, L., {Wu}, Y., and {Wu}, X. (2019).
\newblock Causal modeling-based discrimination discovery and removal: Criteria,
  bounds, and algorithms.
\newblock {\em IEEE Transactions on Knowledge and Data Engineering},
  31(11):2035--2050.

\end{thebibliography}
\bibliographystyle{apalike}

\end{document}